\documentclass[runningheads]{llncs}

\usepackage[mobile]{eccv}

\usepackage{eccvabbrv}

\usepackage{graphicx}
\usepackage{booktabs}

\usepackage[accsupp]{axessibility}

\usepackage[colorlinks=true]{hyperref}

\usepackage{orcidlink}

\renewcommand{\paragraph}[1]{\medskip\noindent\textbf{#1}~}
\renewcommand{\subsubsection}[1]{\paragraph{#1}}

\usepackage{xcolor}
\definecolor{mydarkblue}{rgb}{0,0.08,0.45}
\definecolor{InsetGray}{RGB}{220,220,220}

\colorlet{highlight}{eccvblue}
\usepackage[table]{xcolor}

\usepackage{multirow}
\usepackage{url}
\usepackage{tikz}
\usepackage{adjustbox}

\usepackage{algorithm}
\usepackage{algorithmic}

\usepackage{enumitem}
\usepackage{arydshln}
\usepackage{pifont}
\usetikzlibrary{positioning, calc, shapes.arrows, intersections, arrows.meta, patterns, fadings}
\colorlet{myfinedcolor}{yellow!90!orange}
\colorlet{myfinedbg}{yellow!8}
\usepackage{pgfplots}
\pgfplotsset{compat=1.18}
\usepackage[most]{tcolorbox}
\newtcolorbox{promptbox}[1][]{
  enhanced,
  colback=gray!8,
  colframe=black,
  boxrule=0.8pt,
  sharp corners,
  left=6pt,right=6pt,top=6pt,bottom=6pt,
  width=\linewidth,
  #1
}

\usepackage[capitalize,noabbrev,nameinlink]{cleveref}
\crefname{figure}{Fig.}{Figs.}
\crefname{equation}{Eq.}{Eqs.}
\crefname{appendix}{App.}{Apps.}
\crefname{section}{Sec.}{Secs.}

\newlength{\figurewidth}
\newlength{\figureheight}

\newlength{\imgW}
\newlength{\imgH}
\def\bottomTitlePadX{2mm}
\def\bottomTitlePadY{1.5mm}

\newlength{\leftW}
\newlength{\midW}
\newlength{\rightW}

\newcommand{\ours}{PAWS\xspace}
\newcommand{\bench}{EgoArti\xspace}

\begin{document}

\title{\textcolor{mydarkblue}{\textbf{P}}\textcolor{mydarkblue}{\textbf{A}}\textcolor{mydarkblue}{\textbf{W}}\textcolor{mydarkblue}{\textbf{S}}: \textcolor{mydarkblue}{\textbf{P}}erception of \textcolor{mydarkblue}{\textbf{A}}rticulation in the \textcolor{mydarkblue}{\textbf{W}}ild  \\ at \textcolor{mydarkblue}{\textbf{S}}cale
    from Egocentric Videos}

\titlerunning{Perception of Articulation in the Wild at Scale from Egocentric Videos}

\author{Yihao Wang\textsuperscript{1,6,*}\thanks{$^*$Equal contribution.\quad$^\dagger$Co-advisor.} \and
Yang Miao\textsuperscript{2,*} \and
Wenshuai Zhao\textsuperscript{1,6} \and
Wenyan Yang\textsuperscript{1} \and \\
Zihan Wang\textsuperscript{1} \and
Joni Pajarinen\textsuperscript{1} \and
Luc Van Gool\textsuperscript{2,3} \and
Danda Pani Paudel\textsuperscript{2} \and \\
Juho Kannala\textsuperscript{1,7} \and
Xi Wang\textsuperscript{3,4,5,$\dagger$} \and
Arno Solin\textsuperscript{1,6}}

\authorrunning{Y.~Wang et al.}

\institute{
  $^1$\,Aalto University \quad
  $^2$\,INSAIT, Sofia University \quad
  $^3$\,ETH Zurich \\[2pt]
  $^4$\,TU Munich \quad
  $^5$\,MCML \quad
  $^6$\,ELLIS Institute Finland \quad
  $^7$\,University of Oulu
}

\maketitle

\begin{abstract}
  Articulation perception aims to recover the motion and structure of articulated objects (\eg, drawers and cupboards), and is fundamental to 3D scene understanding in robotics, simulation, and animation.
  Existing learning-based methods rely heavily on supervised training with high-quality 3D data and manual annotations, limiting scalability and diversity.
  To address this limitation, we propose \ours, a method that directly extracts object articulations from hand–object interactions in large-scale in-the-wild egocentric videos.
  We evaluate our method on the public data sets, including HD-EPIC and Arti4D data sets, achieving significant improvements over baselines.
  We further demonstrate that the extracted articulations benefit downstream tasks, including fine-tuning 3D articulation prediction models and enabling robot manipulation. See the project website at \url{https://aaltoml.github.io/PAWS/}.

\end{abstract}

\begin{figure}[h!]
    \centering
    \vspace*{-1.2em}
    \resizebox{\textwidth}{!}{\input{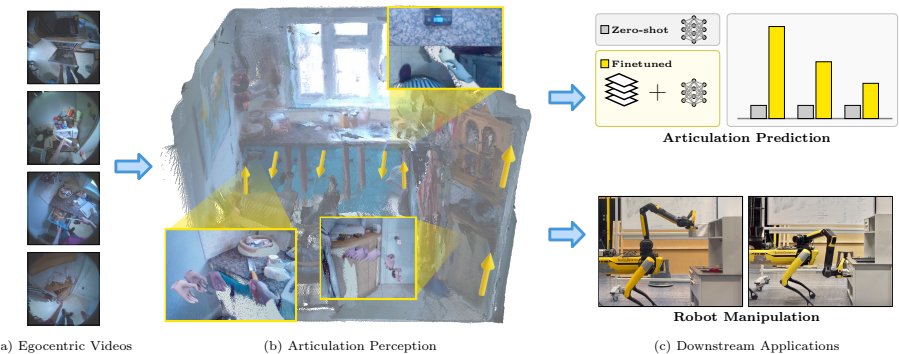}}
    \caption{\textbf{\ours}: Articulation perception and localization from in-the-wild egocentric videos. (a)~From raw videos of human interactions, (b)~our method reconstructs the 3D scene and object articulations using hand cues, geometric recovery, and VLM reasoning. (c)~These serve as annotations to improve downstream articulation prediction models via finetuning, while also providing 3D priors for real-world robotic manipulation.
    }
    \label{fig:teaser}
    \vspace*{-0.5em}
\end{figure}

\section{Introduction}
\label{sec:intro}
Interactions with articulated objects occur everywhere in daily life, from opening a fridge to get breakfast to opening a laptop in the office, and closing a bedside drawer before sleeping.
While current scene understanding enables functional object segmentation \cite{Peng2023OpenScene, corsetti2025fun3du} and scene graph construction \cite{gu_conceptgraphs_2024, zhang_open-vocabulary_2025}, it remains insufficient for agents to reliably interact with articulated parts in the scene (\eg, doors, drawers, and lids) and their interaction interfaces (\eg, knobs and handles).
Studying scene-level articulation is therefore a crucial component of scene understanding, with broad applications in 3D animation and simulation\cite{Object_Wakeup_ECCV2022}, object generation \cite{singapo, dreamart, chen2024urdformer, huang2025react3drecoveringarticulationsinteractive, le2024articulate}, and robotic manipulation \cite{bahl2023affordances, kim_openvla_2024, beingbeyond2025beingh0}.

Recently, learning-based articulation prediction \cite{jiang2022opd, jiayi2023paris, sun2023opdmulti} and generation-based \cite{li2025art} methods have been proposed.
These approaches are typically trained on data sets annotated with articulation information, requiring large-scale labeled data to generalize to unseen objects and scenes.
However, collecting and manually annotating such data sets is costly and labor-intensive.
While several object-level articulation data sets have been introduced \cite{li2019category, geng2022gapartnet}, scene-level data sets remain limited in scale\cite{mao2022multiscan, delitzas_scenefun3d_2024, halacheva2024articulate3d}.
Other methods infer articulation from multi-view observations or temporal data, such as capturing start and end states from image pairs or 3D structure \cite{jiang2022ditto, jiayi2023paris}, or leveraging demonstration videos containing articulated motion \cite{Qian22, kerr2024rsrd, yuan2024general, buchanan2024online, arti25werby}.
Most of these methods still rely on carefully collected data with multi-stage articulation observations, stable lighting conditions, limited occlusions, or additional sensing, such as depth \cite{arti25werby}, significantly hindering scalability and real-world deployment.

In contrast, in-the-wild egocentric videos \cite{grauman2022ego4d, grauman2024ego, EPICFields2023, perrett_hd-epic_2025} are highly diverse, large-scale, and naturally contain rich real human-object interaction information and object kinematic structure.
Despite these advantages, leveraging in-the-wild egocentric videos for articulation understanding remains non-trivial.
These videos are monocular, noisy, sometimes poorly illuminated, and unstructured, making conventional multi-view or temporal methods inapplicable.
We address these limitations by introducing a training-free approach, \ours, that infers scene-level articulation directly from in-the-wild monocular egocentric videos.
Our key insight is that hand-object interactions provide strong physical cues.
By integrating estimated 3D hand trajectories with reconstructed scene geometry, we recover plausible articulation motion and kinematic structure in real-world scenes.
We further leverage foundation-model priors to improve robustness under the noise and occlusions, and validate the resulting annotations through extensive experiments and downstream robotic evaluation.

\subsubsection{Our main contributions} are as follows:
\begin{itemize}[leftmargin=*,noitemsep,topsep=0pt]
\item We introduce a fully automatic, training-free pipeline for scene-level articulation perception from in-the-wild monocular RGB egocentric videos.
\item We propose a foundation-model-driven approach that combines VLMs and LLMs priors to infer plausible articulation motion and structure without task-specific training.
\item We demonstrate through extensive experiments that our method consistently outperforms strong baselines, including Articulation3D and ArtiPoint.
\item We validate the effectiveness and generalizability of the proposed pipeline across downstream 3D articulation prediction and real-world robotic manipulation tasks.
\end{itemize}

\section{Related Work}

\textbf{Articulation Modeling from Static Input} This research direction has seen growing research interest in recent years, since static observations comprise a substantial fraction of real-world visual data.
Existing works have explored articulation prediction from 3D geometry, taking static point clouds as input and jointly predicting part segmentation and motion parameters at either the object level~\cite{wang2019shape2motion, li2019category, yu2024gamma, Deng2023banana, liu2023semi, liu2023selfsupervisedcategorylevelarticulatedobject} or the scene level~\cite{halacheva2024articulate3d}.
~\cite{jiang2022opd, sun2023opdmulti} infer articulation directly from a single RGB(D) image by detecting openable parts and estimating motion parameters.
A separate line of research investigates the generation of articulated 3D objects from image(s)~\cite{singapo, dreamart, physX3D, le2024articulate}.
For instance, PhysX-3D~\cite{physX3D} extends diffusion-based 3D generation frameworks~\cite{trellis} to produce simulation-ready articulated assets from a single image.
While effective, these methods rely on articulation ground-truth annotations and/or high-fidelity 3D structures for training, which are expensive to obtain in unconstrained real-world environments.
The limited availability of such data constrains their scalability and generalization.
More fundamentally, methods based on static inputs inherently lack access to object motion cues and must rely on learned priors or category-level regularities.
As a result, they struggle to reliably recover articulated motion and kinematic structure, particularly in the presence of complex geometries, occlusions, or novel object categories.
In contrast, our approach leverages dynamic observations from in-the-wild egocentric videos, which naturally capture object motion and physical constraints induced by human–object interactions.

\subsubsection{Articulation Modeling from Dynamic Input}
Recent work captures articulations from dynamic inputs using 3D structures~\cite{jiayi2023paris}, RGB images~\cite{Qian22, kim2025screwsplatendtoendmethodarticulated, jain2021icra, jain2022distributional, swaminathan2024leialatentviewinvariantembeddings, shen2025gaussianart, kerr2024rsrd, xu2021d3dhoi, haresh2022articulated, zhang2024artigrasp}, or RGB-D images~\cite{weng2024neural, liu2025building, yu2025artgs3dgaussiansplattinginteractive, buchanan2024online, arti25werby, yuan2024general, dharmarajan2025dream2flow, heppert2022category, gu2025artisgfunctional3dscene}.
RSRD~\cite{kerr2024rsrd} utilizes a hand model to assist articulated object reconstruction, but its reliance on multi-view image inputs is difficult to satisfy in practice.
Tracking-based methods~\cite{yuan2024general, arti25werby, liu2025videoartgsbuildingdigitaltwins, gu2025artisgfunctional3dscene} reconstruct articulation through explicit tracking of openable object parts, which degrades under occlusion, low texture, and motion blur.
Another line of work models human–object interactions~\cite{xu2021d3dhoi, haresh2022articulated, kim2025parahome, nie2022sfa}, leveraging parametric human body models~\cite{SMPL:2015} as constraints for articulation reasoning.
However, these approaches often suffer from reduced reliability due to frequent self-occlusions introduced by the human body during interactions.
In contrast, our method exploits hand information as auxiliary interaction cues, enabling robust 3D articulation perception from unconstrained, in-the-wild egocentric RGB videos.
\cref{tab:prior-comparison} provides a comparison between our approach and prior work.

\newcommand{\rb}[1]{\textbf{#1}}
\begin{table}[t!]
    \centering\scriptsize
    \caption{\textbf{Comparison to prior art.}
    We summarize related articulation perception methods by input modality,
    articulation perception strategy (SL: supervised learning; SSL: self-supervised learning),
    supervision format, and output format. \emph{PC} denotes point clouds, which correspond to
    high-fidelity scanned meshes in~\cite{halacheva2024articulate3d}.
    }
    \vspace*{-0.1in}
    \setlength{\tabcolsep}{16pt}
    \begin{tabular}{l c c c}
        \toprule
        Method & \rb{Input} & \rb{Supervision} & \rb{Output} \\
        \midrule
        Articulation3D~\cite{Qian22} & RGBs & SL, 3D anno. & 3D planes \\
        OPDMulti~\cite{sun2023opdmulti} & RGB & SL, 2D masks & 2D masks \\
        USDNet~\cite{halacheva2024articulate3d} & PC & SL, 3D anno. & PC \\
        RSRD~\cite{kerr2024rsrd} & RGBs & SSL, RGBs & Meshes \\
        Articulate-Anything~\cite{le2024articulate} & RGBs & Transfer Learning & Meshes\\
        iTACO~\cite{peng_itaco_2025} & RGB-Ds & Transfer Learning & Meshes\\
        ArtiPoint~\cite{arti25werby} & RGB-Ds & Transfer Learning & 3D joints \\
        \rowcolor{highlight!10!white}
        \ours\ (ours) & RGBs & Transfer Learning & 3D joints \\
        \bottomrule
    \end{tabular}
    \label{tab:prior-comparison}
    \vspace{-1em}
\end{table}

\subsubsection{Hand/Hand-Object Interaction Reconstruction}
Early work in this area primarily focused on hand detection \cite{Fouhey_2018_CVPR}, segmentation\cite{narasim2019hand}, or contact state recognition from 2D images~\cite{Shan20, zhang_fine-grained_2022}.
Cheng et al.~\cite{cheng_towards_2023} extended this line of research to multi-object contact segmentation.
Recently, ~\cite{MANO:SIGGRAPHASIA:2017} have investigated hand-object reconstruction in 3D, typically representing the hand using the MANO hand model.
Hand meshes can be estimated from a single image~\cite{pavlakos2024reconstructing, potamias_wilor_2025} or from consecutive video sequences~\cite{zhang2025hawor}.
There has also been work on joint hand-object reconstruction in 3D, where objects are represented either as meshes~\cite{swamy2025host3rkeypointfreehandobject3d, ye2025whole} or as point clouds~\cite{chen_hort_2025}.
In addition, several recent approaches leverage hand information for zero-shot transfer from human demonstrations to robot control~\cite{yuan2024general, chen2025vidbot, yang2025egovla, beingbeyond2025beingh0, kuang2024ram}.
Our task focuses on reconstructing object articulation from RGB videos of natural interactions involving both hands and articulated furniture.

\subsubsection{Articulated Objects Data sets}
Prior work has focused on object-level motion annotations and part mobility benchmarks~\cite{wang2019shape2motion, liu2022akb, Mo_2019_CVPR, geng2022gapartnet}. Scene-level data sets capture articulated environments with rich spatial context~\cite{mao2022multiscan, sun2023opdmulti, halacheva2024articulate3d, delitzas_scenefun3d_2024}.
Some recent efforts also incorporate human-object interaction data with object and articulation labels~\cite{xu2021d3dhoi, kim2025parahome, arti25werby}.
Despite recent progress, the overall volume of articulated data remains limited, as acquiring detailed part geometry (\eg, via RGB-D scanning or 3D mesh reconstruction) and annotating articulation parameters are both costly and labor-intensive.
In contrast to existing data sets, which mostly require controlled sensors or structured environments, our approach recovers 3D articulation directly from monocular egocentric RGB videos.
This formulation enables scalable articulation annotation from large-scale, unstructured real-world data without requiring curated 3D scans or explicit 3D supervision.

\begin{figure*}[h!]
    \centering
    \includegraphics[width=\linewidth]{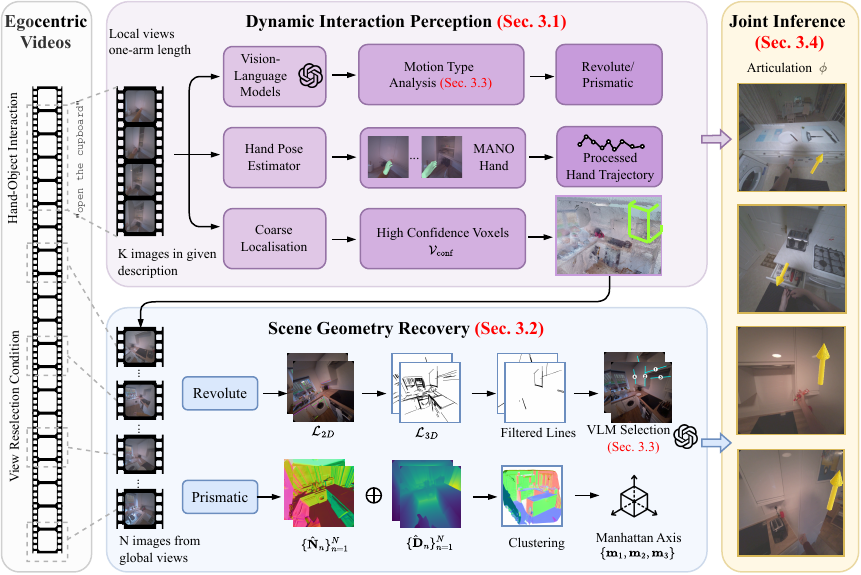}
    \caption{\textbf{Overall pipeline.} Given a full in-the-wild egocentric video and a language description as input, our pipeline consists of four parts:
\textbf{(1) Dynamic Interaction Perception:} We first segment the video based on the language description and extract interactive frames (referred to as "local views"), 3D hand trajectories, motion types, and coarse object localizations.
\textbf{(2) Geometric Structure Recovery:} Based on the object's location, we select "global views" from the full video. Depending on the motion type, we recover the scene geometry using different flows.
\textbf{(3) VLM-guided Reasoning:} The VLM first infers the motion type to provide a prior for global view selection, and then identifies plausible articulation axes during the geometry recovery stage.
\textbf{(4) Joint Articulation Inference:} We integrate 3D hand trajectories and the recovered geometry to infer the final articulations.
    }
    \label{fig:pipeline}
\end{figure*}

\section{Method}
\label{sec:method}
\textbf{Problem Formulation}
An articulated object consists of multiple rigid parts interconnected by joints. Both organic (\eg, the human body) and human-made objects (\eg, drawers) can exhibit articulation \cite{liu2025survey}.
In this work, we focus on the articulation of human-made objects and simplify each joint as having one degree of freedom (DoF).
Following previous benchmarks \cite{mao2022multiscan, delitzas_scenefun3d_2024, halacheva2024articulate3d}, we represent each articulated joint $\phi_i$ by three motion parameters: motion type $c_i$, motion axis $a_i$, and motion origin $o_i$
\begin{equation}
\phi_i = \{c_i, a_i, o_i\}.
\end{equation}
The motion type is defined as $c_i \in \{\text{prismatic}, \text{revolute}\}$.
The motion axis describes the rotation axis for revolute motion or the translation direction for prismatic motion, denoted as $a_i \in \mathbb{R}^3$, while the motion origin $o_i \in \mathbb{R}^3$ defines the pivot point for revolute motion.

\subsubsection{Objective} We aim to estimate the articulation parameters $\phi$ from untrimmed, in-the-wild egocentric RGB videos $\mathbf{V} \in \mathbb{R}^{T\times H\times W\times 3}$ and a set of action descriptions $\mathcal{L} = \{ l_1, l_2, \dots, l_M \}$ across the whole video. Here, $l_j \in \mathcal{L}$ represents a specific action description (\eg, open the door'', close the fridge'') and its corresponding temporal boundaries $\tau_{\text{start}}^{(j)}:\tau_{\text{end}}^{(j)}$ in the video.
The language descriptions can be obtained either from manual annotations or from grounding model inferences.The articulation estimation includes the object-level articulation $\phi_{obj} = \Phi(\mathbf{V}, l_j)$ and the aggregated scene-level articulation $\phi_{scene} = \Phi(\mathbf{V}, \mathcal{L})$.
The operator $\Phi(\cdot)$ encapsulates the overall articulation estimation process.

\subsubsection{Pipeline}
Our framework is illustrated in \cref{fig:pipeline}. It infers articulation types and parameters through the following core components:
(1) The \textit{Dynamic Interaction Perception} module utilizes specific interaction sequences to provide 3D hand motion cues, coarse interacted object locations, and motion types, which are essential for anchoring the motion origin $o_i$ and subsequent geometry recovery (\cref{sec:method:hand});
(2) The \textit{Static Scene Geometry Recovery} module reconstructs the scene structure and generates candidate articulation axes, serving as geometric priors for $a_i$ (\cref{sec:method:geo});
(3) The \textit{VLM-Aided Motion Reasoning} stage infers the motion type $c_i$ and selects physically plausible revolute axes from the geometric candidates (\cref{sec:method:vlm}); and
(4) The \textit{Joint Articulation Inference} process integrates all acquired information to output the final articulation parameters (\cref{sec:method:final}).
We emphasize that the VLM is employed to assist in dynamic interaction and static structure analysis as an integrated reasoning component, rather than being treated as a decoupled module.
By design, our pipeline is robust to scale variations and uses a single set of hyperparameters across all data sets. Implementation details and hyperparameter settings are provided in \cref{supp:impl}.

\begin{algorithm}[t]
\footnotesize
\renewcommand{\algorithmiccomment}[1]{\hfill $\triangleright$ #1}
\caption{Static Geometry Recovery Pipeline}
\begin{algorithmic}[1]
\REQUIRE Video frames $\mathbf{V}$, camera poses $\mathbf{P}$, intrinsics $\mathbf{K}$, number of sampled views $N$

\STATE $\mathcal{I}_s \leftarrow \text{SelectViews}(\mathbf{V}, \mathbf{P}, N)$ \COMMENT{Sample representative views}

\FOR{each view $I_n \in \mathcal{I}_s$}
    \STATE $(\mathbf{D}_n, \mathbf{N}_n) \leftarrow \text{MoGe}(I_n)$ \COMMENT{Estimate metric depth and surface normals}
\ENDFOR

\STATE $\mathcal{L}_{3D} \leftarrow \emptyset$ \COMMENT{Triangulated 3D lines}

\STATE $\mathcal{L}_{2D} \leftarrow \text{DetectLines}(\mathcal{I}_s)$

\FOR{each line correspondence across views}
    \STATE $\mathbf{P}_{3D} \leftarrow \text{Triangulate}(\mathcal{L}_{2D}, \{\mathbf{D}_n\}, \mathbf{K}, \mathbf{P})$
    \STATE $\mathcal{L}_{3D} \leftarrow \mathcal{L}_{3D} \cup \mathbf{P}_{3D}$
\ENDFOR

\STATE \tikz[baseline=-0.5ex]{\draw[dashed, dash pattern=on 4pt off 2pt, line width=0.5pt, gray] (0,0) -- (0.92\linewidth,0);}
\addtocounter{ALC@line}{-1}

\STATE $\mathcal{C} \leftarrow \text{ClusterDirections}(\mathcal{L}_{3D})$

\FOR{each cluster $c \in \mathcal{C}$}
    \STATE $\mathbf{a}_c \leftarrow \text{LO-RANSAC}(\text{lines in } c)$ \COMMENT{Revolute motion axis candidates}
\ENDFOR

\STATE $\mathcal{N} \leftarrow \bigcup_{I_n \in \mathcal{I}_s} \text{SampleNormals}(\mathbf{N}_n)$

\STATE $\mathbf{M} \leftarrow \text{ClusterNormals}(\mathcal{N})$

\STATE $\{\mathbf{m}_1,\mathbf{m}_2,\mathbf{m}_3\} \leftarrow \text{ManhattanAxes}(\mathbf{M})$ \COMMENT{Prismatic motion axis candidates}

\end{algorithmic}
\label{alg:geometry}
\end{algorithm}

\subsection{Dynamic Interaction Perception}
\label{sec:method:hand}

Given the complete video sequence $\mathbf{V}$ and a language description $l_j$ of a single action, we first segment the temporal sub-sequence $\mathbf{v}_j = \mathbf{V}[\tau_{\text{start}}^{(j)}:\tau_{\text{end}}^{(j)}] $ of the interaction based on its temporal boundaries.
We denote that there are $K$ RGB frames inside this segment $\mathbf{v}_j$, and we obtain the dynamic hand trajectory and the object's coarse localization from them.

\subsubsection{Metric-Aware 3D Hand Reconstruction}
We first apply a frame-based 3D hand reconstruction method~\cite{potamias_wilor_2025, zhang2025hawor} to identify frames containing visible hands and to reconstruct the MANO~\cite{MANO:SIGGRAPHASIA:2017} hand pose for each selected frame.
Hands may be occluded or move out of view in certain frames; thus, we obtain $K'$ hand poses, where $K'\le K$.
According to grasp statistics from \cite{Brahmbhatt_2020_ECCV, GRAB:2020}, we use the thumb, index, and middle fingertip landmarks across the $K'$ frames to represent the hand trajectories $\{\mathbf{z}_t \in \mathbb{R}^3\}_{t=1}^{K'}$ at observed timestamps $\{t\}_{t=1}^{K'}$.
Unlike prior approaches that lift 2D hand keypoints to 3D using relative depth information (\eg,~\cite{chen2025vidbot}), our method directly estimates the 3D hand pose and thus avoids explicit reliance on depth scaling.

\subsubsection{Stochastic Hand Trajectories Refinement}
The raw hand trajectory $\{\mathbf{z}_t\}$ across a video is typically noisy, as it may include hand motions before, during, and after object contact, alongside errors introduced by the hand pose estimation model.
In addition, unconstrained human motion introduces further uncertainty.
During the active interaction phase, the hand theoretically remains in contact with the object surface, physically bound to a rigid kinematic manifold (\eg, a straight line or a circular arc).
Human motor control in such constrained manipulation tasks naturally exhibits mean-reverting properties.
Therefore, we integrate Ornstein-Uhlenbeck (OU) approach and Rauch--Tung--Striebel (RTS) smoothing to obtain the articulation-related trajectories $\{\hat{\mathbf{p}}_t \in \mathbb{R}^3\}$, which are crucial for determining the final motion origin $o_i$.
Please refer to \cref{supp:method:hand} for more details.

\subsubsection{Object Coarse Localization}
We first estimate a coarse geometric proxy of the indoor scene to localize the interacted object.
We use 3D reconstruction approach \cite{keetha2026mapanything} reconstruct a coarse scene representation from the interaction clip $\mathbf{v}_j$ and discretize the resulting point cloud into a 3D voxel grid $\mathcal{V}$.
Each voxel $v \in \mathcal{V}$ is associated with an observation count $n_v$, defined as the number of camera views in which the voxel is consistently observed.
A subset of high-confidence voxels is then extracted as $\mathcal{V}_{\text{conf}} = \{v \in \mathcal{V} \mid n_v > T_{\mathrm{conf}}\}$, which corresponds to spatial regions that are likely to belong to the manipulated object.
This coarse localization step provides a robust spatial prior $\mathcal{V}_{\text{conf}}$ that is resilient to occlusions and hand-induced motion.

\subsection{Static Scene Geometry Recovery}
\label{sec:method:geo}

Egocentric videos typically exhibit limited viewpoint variation, as the camera motion is constrained to within arm’s length during human–object interactions.
At the same time, frequent hand motion and self-occlusions significantly complicate the geometric reconstruction of articulated objects.
As a result, directly recovering geometry from per-frame depth estimates or 2D tracking alone \cite{arti25werby, liu2025videoartgsbuildingdigitaltwins} often leads to unstable and inconsistent results in in-the-wild scenarios.
To address these challenges, \ours~adopts a coarse-to-fine geometric recovery strategy that integrates camera view reselection and multi-view geometry reconstruction.
Following \cite{halacheva2024articulate3d}, we utilize motion type priors and reconstruct revolute and prismatic candidate axes $a_i$ using different geometric strategies.
The full algorithmic details of geometry recovery are provided in \cref{alg:geometry}.

\subsubsection{Frustum Intersection and View Reselection}Given the coarse object localization represented by the high-confidence voxel region $\mathcal{V}_{\text{conf}}$, we sample a subset of $N$ global camera views $\mathcal{I}_s = \{I_1, \dots, I_N\}$ from the full egocentric video for multi-view geometry reconstruction.
We first retain frames whose camera frustums intersect $\mathcal{V}_{\text{conf}}$.Among these candidates, we apply farthest point sampling over camera positions to select a subset of views that maximizes spatial coverage.
Finally, we refine the selected views using object-specific semantic masks~\cite{liu2023grounding, ren2024grounded, kirillov2023segany} $\mathcal{M} \in \{0, 1\}^{N \times H \times W}$, segmented using the object description $l_j$ or a foundation model, ensuring consistent object localization across views.

\subsubsection{Multi-View Line Triangulation for Revolute Priors}
To generate candidate axis proposals for revolute motion, we draw inspiration from the line matching paradigm in~\cite{Liu_2023_LIMAP}.
For each selected camera view $I_n \in \mathcal{I}_s$, we first detect a set of 2D line segments $\mathcal{L}_{2D}$ using~\cite{Pautrat_2023_DeepLSD}.
For each detected 2D line segment $l_{2D} \in \mathcal{L}_{2D}$, we densely sample a set of pixel coordinates $\mathcal{S}_{l_{2D}} = \{\mathbf{u}_m = (u_m, v_m)\}$.
To avoid the scale ambiguity inherent to traditional monocular depth estimation, we estimate a metric-scale depth map $\mathbf{D}_n$ using a monocular geometry model~\cite{wang2025moge2}.
Instead of performing explicit multi-view epipolar triangulation, the sampled 2D pixels $\mathbf{u}_m$ are directly unprojected to 3D camera coordinates using the camera intrinsic matrix $\mathbf{K}$ and their corresponding metric depth $\mathbf{D}_n(\mathbf{u}_m)$.
These points are subsequently transformed into world coordinates using the camera-to-world pose $\mathbf{P}_n$ to form a 3D point set $\mathcal{P}_{l_{2D}}$.
By aggregating the metric 3D point sets $\mathcal{P}_{l_{2D}}$ across multiple views based on 2D line correspondences, we obtain dense 3D point tracks.
Finally, we apply LO-RANSAC to robustly fit a 3D line model—parameterized by a direction vector and an origin point—to these tracks, generating the discrete set of candidate revolute axis proposals $\mathcal{L}_{3D} = \{l_{3D}^{(1)}, l_{3D}^{(2)}, \dots \}$.

\subsubsection{Manhattan Spatial Priors for Prismatic Motion}
To better describe prismatic motion, we adopt the Manhattan-world assumption~\cite{coughlan2000manhattan}.
We estimate the Manhattan coordinate system $\{\mathbf{m}_1, \mathbf{m}_2, \mathbf{m}_3\}$ using the metric depth maps $\{\mathbf{D}_n\}_{n=1}^N$ and surface normal maps $\{\mathbf{N}_n\}_{n=1}^N$ predicted by the geometry estimation model~\cite{wang2025moge2}.
Specifically, we aggregate and cluster the surface normals $\mathcal{N} = \bigcup_{n=1}^N \mathbf{N}_n$ across the $N$ sampled images to extract the dominant orthogonal directions.
These vectors serve as our candidate set for prismatic motion axes. Detailed descriptions are provided in \cref{supp:manhattan}.

\subsection{Vision-Language Guided Motion Reasoning}
\label{sec:method:vlm}

\begin{figure*}[t!]
    \centering
    \resizebox{\textwidth}{!}{\includegraphics{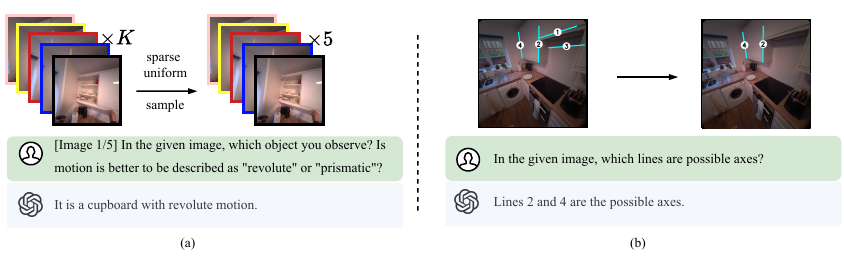}}
    \caption{\textbf{Illustration of VLM Reasoning.}
    \textbf{(a)} Temporal Motion Type Classification.
    \textbf{(b)} Spatial Axis Grounding via Set-of-Marks VQA.
    }
    \label{fig:method:vlm}
\end{figure*}

While humans intuitively infer motion types and articulation axes from visual cues, this remains a significant challenge for vision-based algorithms due to noisy hand trajectories and dense geometric candidates. To address this, we leverage a Vision-Language Model (VLM) through a two-stage Visual Question Answering (VQA) pipeline, with each stage targeting a distinct kinematic reasoning objective.

\subsubsection{Temporal Motion Type Classification}
The first stage classifies the motion type of the articulated joint $i$ as either \textit{revolute} or \textit{prismatic}, denoted as $c_i \in \{\text{revolute}, \text{prismatic}\}$.
In-the-wild hand trajectories are inherently noisy; for example, after pulling a drawer, a user’s hand may slide across the surface, creating geometric ambiguity.
To provide coarse temporal context while maintaining efficiency, we follow a sparse sampling strategy.
Given the interaction clip $\mathbf{v}_j$, we uniformly sample $K$ frames $\{f_1, f_2, \dots, f_K\}$ at fixed temporal indices.
For each sampled frame, the VLM is queried to identify the interacted object and its frame-level motion type.
These per-frame predictions are then aggregated by an LLM to produce a robust global motion type decision $c_i$, effectively filtering out transient noise or post-interaction artifacts.

\subsubsection{Spatial Axis Grounding via Set-of-Marks VQA}
In the second stage, we identify the most plausible articulation axis from the set of reconstructed 3D line candidates $\mathcal{L}_{3D}$.
While our geometric module proposes candidates within the Manhattan coordinate system, many represent irrelevant static structures, such as table corners or appliance edges.
To bridge the gap between continuous 3D space and VLM reasoning, we reformulate axis selection as a constrained multiple-choice VQA problem.
For a selected camera view, we project the 3D line candidates back into the 2D image plane.
To ensure relevance, we filter these projections using the semantic mask $\mathcal{M}$ generated by the grounding model, such that $\mathcal{L}_{\text{filtered}} = \{l_{3D} \in \mathcal{L}_{3D} \mid \text{proj}(l_{3D}) \cap \mathcal{M} \neq \emptyset\}$.
From this subset, we select the $N_{\text{cand}}$ most prominent candidates (\eg, $N_{\text{cand}}=4$ based on segment length) and overlay them as numbered visual prompts (Set-of-Marks).
The VLM is prompted to select the index $k$ that best represents the semantic articulation hinge based on physical commonsense.
This yields a semantically verified candidate subset $\mathcal{L}_{3D}^* \subset \mathcal{L}_{3D}$ that guides the final multi-view geometric consistency refinement.

\subsection{Joint Articulation Inference}
\label{sec:method:final}
After inferring the motion type $c_i \in \{\text{prismatic}, \text{revolute}\}$ via VLM reasoning, we estimate the final articulation parameters by jointly leveraging the dynamic hand-object interaction information from \cref{sec:method:hand} and the static geometric structure \cref{sec:method:geo}.
Articulation can be inferred from a single interaction clip or aggregated across multiple clips within the same scene. The PCA and RANSAC are used to obtain final the robust articulation parameters $\phi_i = \{ c_i, a_i, o_i \}$
For more details, please refer to \cref{supp:method:final}.

\section{Experiments}

We evaluate the effectiveness and versatility of our articulation perception pipeline through a series of experiments. Specifically, we demonstrate that:
(1) Our method significantly outperforms strong baselines on the articulation perception task using in-the-wild egocentric RGB videos.
(2) Key components of our pipeline, including hand-object interaction cues and VLM guidance, are critical to achieving strong performance.
(3) The articulation labels extracted by our pipeline improve the performance of existing articulation prediction models through downstream fine-tuning.
(4) The recovered articulation parameters can be directly applied to real-world robotic manipulation tasks.

\subsection{Accuracy of Articulation Perception}
\textbf{Data Sets}
To evaluate our method, we utilize the HD-EPIC~\cite{perrett_hd-epic_2025}, Arti4D~\cite{arti25werby}, and Epic-Fields~\cite{EPICFields2023} data sets, where videos are segmented into temporal clips based on provided language descriptions.
The original videos span several minutes, and for the HD-EPIC and Epic-Fields data sets, they include a vast range of background actions unrelated to our task (\eg, washing hands or holding small rigid objects).
To ensure a focused evaluation, we curate a specific subset by selecting clips where the language descriptions explicitly indicate hand-object interactions with articulated furniture.
This process yields 313 annotated interaction clips involving 80 furniture instances across 9 scenes for HD-EPIC, and over 6,000 selected clips spanning 34 scenes for Epic-Fields.
Because Arti4D is natively designed for scene-level articulated interactions, we utilize its standard evaluation splits without additional filtering.
While Arti4D and HD-EPIC serve as the basis for both our quantitative and qualitative analyses, we utilize Epic-Fields primarily for large-scale qualitative evaluation.

\subsubsection{Baselines}
We compare our approach against Articulation3D~\cite{Qian22}, Articulate-Anything~\cite{le2024articulate} (AA), ArtiPoint~\cite{arti25werby}, and iTACO~\cite{peng_itaco_2025}.
We reimplement two Articulation3D-based variants. The original Articulation3D assumes a static camera and consistent object motion, assumptions that are severely violated in in-the-wild egocentric settings. \sloppy
Therefore, we incorporate moving camera poses into the temporal optimization, denoted as Articulation3D$^{*}$.
We further relax the regression-based temporal filtering, which assumes constant motion velocity, resulting in Articulation3D$^{**}$.
For both variants, we use the per-frame predicted 2D masks and articulation axes provided by Articulation3D, lifting them to 3D using the predicted metric depth from the monocular geometry model~\cite{wang2025moge2}.
For ArtiPoint~\cite{arti25werby} and iTACO~\cite{peng_itaco_2025}, we use the ground-truth camera poses from the data sets instead of those estimated by visual odometry~\cite{teed2021droid, peng_itaco_2025}.
This is to isolate and evaluate the pure performance of the tracking-based articulation extraction and ensure a fair comparison.
Regarding depth inputs for 3D lifting, we use predicted metric depth~\cite{wang2025moge2} for experiments on HD-EPIC~\cite{perrett_hd-epic_2025}, and utilize the natively provided depth sensor information for experiments on Epic-Fields~\cite{EPICFields2023} and Arti4D~\cite{arti25werby}.
In this part, we focus on the results of the experiment on Articulation3D and ArtiPoint.
Further implementation specifics and experiment results are provided in \cref{supp:impl}.

\subsubsection{Metrics}
We adopt the articulation motion evaluation metrics from prior work~\cite{jiang2022opd, sun2023opdmulti, delitzas_scenefun3d_2024, halacheva2024articulate3d}.
Each predicted articulation instance is matched to its corresponding ground-truth instance.
To quantify articulation perception accuracy, we report the average accuracy corresponding to the components of our estimated articulation formulation $\phi_i = \{c_i, a_i, o_i\}$ under three criteria:
motion type $c_i$ ($\mathbf{M}$),
motion type combined with motion axis $a_i$ ($\mathbf{MA}$), and
motion type combined with both motion axis and motion origin $o_i$ ($\mathbf{MAO}$).
Because articulation recognition may inherently fail due to upstream errors in hand–object interaction detection or object tracking, we report results under two evaluation settings to comprehensively reflect inference capability:
(i) conditioned solely on successfully detected articulation instances, and
(ii) unconditionally over all video clips, thereby penalizing missed detections.
All results are averaged across scenes.

Consistent with our kinematic formulation, the motion origin $o_i$ is evaluated exclusively for revolute motions.
Following standard protocols, an axis prediction is considered correct if the angular error is within $15^\circ$ (evaluated via a cosine distance threshold of $1 - \cos(15^\circ)$), and an origin prediction is considered correct if the Euclidean distance error is within $0.25 \mathrm{m}$.
Note that while existing static 3D benchmarks typically evaluate articulation based on the volumetric IoU of predicted openable regions, our task focuses dynamically on articulation extraction from unconstrained egocentric videos; therefore, we evaluate these metrics strictly on a per-interaction basis.

\begin{table*}[t]
\centering
\caption{\textbf{Quantitative results on articulation estimation accuracy.}
Our method achieves higher accuracy on the HD-Epic~\cite{perrett_hd-epic_2025} and Arti4D~\cite{arti25werby} data sets, outperforming the RGB-D–based articulation extraction baseline.
Note that in the HD-Epic data set, since not every articulable fixture involves hand interaction, we evaluate only the actually interactive fixtures. The evaluated pairs means the actual evaluated articulation.
We denote \ours-s as aggregated articulation estimates at the scene level.
Since the scene configuration varies across videos in the Arti4D data set, we do not report aggregated results for this data set.}
\label{tab:quantitative_1}
\vspace*{-0.1in}
\resizebox{\textwidth}{!}{
\begin{tabular}{lcccccccccccccc}
\toprule
& \multicolumn{7}{c}{\textbf{HD-EPIC} \cite{perrett_hd-epic_2025}}
& \multicolumn{7}{c}{\textbf{Arti4D} \cite{arti25werby}} \\
\cmidrule(lr){2-8} \cmidrule(lr){9-15}

\textbf{Methods}
& Match ($\%$)
& M
& MA
& MAO
& M$^\dagger$
& MA$^\dagger$
& MAO$^\dagger$
& Match ($\%$)
& M
& MA
& MAO
& M$^\dagger$
& MA$^\dagger$
& MAO$^\dagger$ \\
\midrule

Articulation3D$^{*}$~\cite{Qian22}
& 22.68 & 0.24 & 0.12 & 0.08 & 0.95 & 0.48 & 0.35 & 45.07 &
0.32 & 0.03 & 0.02 & 0.70 & 0.07 & 0.05 \\

Articulation3D$^{**}$~\cite{Qian22} & 42.22 & 0.34 & 0.15 & 0.11 & 0.81 & 0.37 & 0.27 & 84.29 & 0.53 & 0.06 & 0.03 & 0.62 & 0.07 & 0.04\\

\cdashline{2-15}
ArtiPoint~\cite{arti25werby}
& 70.66 & 0.47 & 0.06 & 0.00 & 0.72 & 0.09 & 0.01
& \textbf{85.02} & \textbf{0.63} & \textbf{0.52} & \textbf{0.47} & 0.75 & 0.61 & 0.56 \\
\cdashline{2-15}
\ours
& 55.38 & 0.52 & 0.36 & 0.20 & 0.96 & \textbf{0.66} & 0.35 & 48.02
& 0.47 & 0.37 & 0.34 & \textbf{1.00} & \textbf{0.80} & \textbf{0.75} \\

\ours-s
& \textbf{71.43} & \textbf{0.71} & \textbf{0.46} & \textbf{0.36} & \textbf{0.98} & 0.63 & \textbf{0.48}
& -- & -- & -- & -- & -- & -- & -- \\

\bottomrule
\end{tabular}
}
\end{table*}

\subsubsection{Comparison on HD-EPIC Data Set \cite{perrett_hd-epic_2025}}
As shown in \cref{tab:quantitative_1}, our method consistently outperforms all baselines on the HD-EPIC~\cite{perrett_hd-epic_2025} data set.
We observe that Articulation3D succeeds on only a small subset of clips and fails on most in-the-wild egocentric videos.
A key reason is the domain gap: Articulation3D is pre-trained on Internet videos with relatively sufficient viewpoint coverage, whereas egocentric videos often contain incomplete observations of the articulated object and rapidly changing viewpoints.
These factors degrade both per-frame openable-part prediction and subsequent temporal optimization.
The higher accuracy of Articulation3D$^{**}$ further illustrates that its assumption of consistent object motion is frequently violated in real-world egocentric interactions, where objects may move at varying speeds or even reverse direction.

ArtiPoint~\cite{arti25werby} performs well when hand–object interactions are clearly observable and tracking succeeds.
However, actions in in-the-wild data sets are often fast, resulting in a limited number of usable frames for reliable tracking.
Moreover, rapid viewpoint changes and low-texture furniture surfaces (\eg, white cupboards) further increase tracking difficulty.
Finally, monocular depth prediction errors from~\cite{wang2025moge2} introduce additional bias during 3D lifting on HD-EPIC~\cite{perrett_hd-epic_2025}.
In contrast, our method leverages hand estimation and global geometric constraints to robustly address these challenges in in-the-wild egocentric settings.

\subsubsection{Comparison on Arti4D Data Set \cite{arti25werby}.}
On the Arti4D data set, our approach achieves performance comparable to that on HD-EPIC.
However, we observe that ArtiPoint performs substantially better on Arti4D than on HD-EPIC.
This improvement can be attributed to the characteristics of the Arti4D data set, which contains more textured furniture surfaces and visual markers, as well as more stable hand and camera motion during interactions.
Consequently, Arti4D is more suitable for object tracking–based methods than in-the-wild settings.
In addition, the availability of accurate depth information further reduces the uncertainty of 3D lifting for ArtiPoint.
Articulation3D-based baselines also perform better on Arti4D than on HD-EPIC, as failures in openable object detection in in-the-wild scenarios can lead to incorrect object associations and disrupt subsequent optimization.
Despite these differences, among successfully detected articulations, our approach still achieves competitive performance.

\subsection{Ablation Study}
To analyze the contribution of individual components in our pipeline, we conduct a series of ablation experiments.
All ablations are evaluated on the HD-EPIC data set~\cite{perrett_hd-epic_2025} and shown in \cref{tab:ablation}.
While we instantiate our pipeline using \cite{wang2025moge2} for geometry and GPT-4o for semantic reasoning due to their current state-of-the-art performance, our formulation is agnostic to the specific choice of foundational models.
We also provide the robustness analysis for different VLM model usage and other foundation models in \cref{supp:exp}.

\begin{table}[t]
\centering
\caption{\textbf{Ablation study.}
Component-wise ablation demonstrating the necessity of the Manhattan Geometric Filter and the GPT-4o reasoning engine.}
\label{tab:ablation}
\vspace*{-0.1in}
\begin{tabular}{ll | ccc}
\toprule
\textbf{Ablation Type} & \textbf{Model / Configuration Variant} & \textbf{M} & \textbf{MA} & \textbf{MAO} \\
\midrule

\multirow{4}{*}{\shortstack[l]{Architecture\\Components}}
& w/o Filter, w/o GPT-4o (Heuristic) & 0.46 & 0.28 & 0.13 \\
& w/ Filter, w/o GPT-4o              & 0.45 & 0.30 & 0.15 \\
& w/o Filter, w/ GPT-4o              & \textbf{0.47} & 0.32 & \textbf{0.18} \\
& \textbf{Full PAWS (Filter + GPT-4o)} & 0.46 & \textbf{0.33} & \textbf{0.18} \\

\bottomrule
\end{tabular}
\end{table}

\subsubsection{Impact of Hand Contact Constraints}
When hand interaction cues are removed, the method relies solely on VLM-based axis proposals, which are insufficient to reliably determine object articulation.
As shown in \cref{tab:ablation}, incorporating hand trajectory smoothing and contact-based filtering leads to a slightly improvement in accuracy.
This improvement arises because trajectory smoothing reduces noise introduced by the hand motion estimator~\cite{zhang2025hawor}, while the hand contact detector~\cite{cheng_towards_2023} effectively filters out trajectory points that do not correspond to actual object interaction.

\subsubsection{Impact of VLM Reasoning}
Incorporating VLM-based reasoning improves articulation accuracy compared to geometry-only inference.
Specifically, applying the VLM filter increases performance from 0.30 to 0.33 in MA and from 0.15 to 0.18 in MAO (see \cref{tab:ablation}), indicating that language-guided reasoning helps select more plausible articulation axes.
However, VLM-based geometry filtering alone is insufficient to fully determine articulation.
In ambiguous cases, such as low-texture furniture where multiple edges appear visually similar (\eg, white cupboards), VLM predictions may be unreliable, highlighting the necessity of grounding articulation inference in physical hand-object interaction cues.
\Cref{tab:qualitative-finetune} presents the performance gains obtained after fine-tuning with our annotated data.

\begin{table}[t]
\centering\scriptsize
\caption{
\textbf{Quantitative results on articulation prediction using the \bench benchmark.}
The results before and after fine-tuning on our data set demonstrate the effectiveness of our benchmark.
\textbf{MS} refers to the average accuracy of predictions with correct motion type and correct object mask prediction (IoU larger than 50$\%$), similar cases for \textbf{MSA} and \textbf{MSAO}.
We evaluate different fine-tuning strategies across multiple data sets.
}
\label{tab:qualitative-finetune}
\setlength{\tabcolsep}{6pt}
\vspace*{-0.1in}
\begin{tabular}{l|cccc|ccc}
\toprule
& \multicolumn{4}{c}{HD-Epic \cite{perrett_hd-epic_2025}} & \multicolumn{3}{c}{Arti4D \cite{arti25werby}} \\
{\bf Methods} & \bf M  & \bf MS & \bf MSA & \bf MSAO & \bf M & \bf MA & \bf MAO  \\
\midrule
Articulate3D (Zero-shot) & 0.46 & 0.02 & 0.02 & 0.02  & 0.41 & 0.14 & 0.12 \\
Articulate3D (Finetuned) & \textbf{0.58} & \textbf{0.13} & \textbf{0.08} & \textbf{0.05} & \textbf{0.48} & \textbf{0.22} & \textbf{0.17} \\
\bottomrule
\end{tabular}
\end{table}
\vspace*{-0.1in}

\subsection{Downstream Applications}
\textbf{Cross-Data Set Generalization}
We evaluate the effectiveness of \ours~for improving scene-level articulation prediction by finetuning USDNet~\cite{halacheva2024articulate3d} using data automatically extracted by our pipeline.
Specifically, we generate articulation annotations on the HD-Epic \cite{perrett_hd-epic_2025} data set using \ours, and use them to augment the original USDNet training set.
We refer the new data set \bench.
For each scene in \bench, we aggregate articulation predictions across all egocentric trajectories to obtain a unified scene-level articulation representation.
Semantic masks for each articulated object are obtained from the digital twin provided by~\cite{perrett_hd-epic_2025}.
We finetune USDNet on the \bench data set and evaluate the model before and after finetuning on both \bench~(in-domain) and Arti4D data set~(cross-data set).
As shown in \Cref{tab:qualitative-finetune}, the model finetuned with \ours-labeled data significantly outperforms the zero-shot baseline across both data sets, demonstrating that our automatic labeling pipeline produces high-quality supervision that improves in-domain performance and cross-data set generalization.

\subsubsection{Robotic Manipulation}
\begin{figure}[t!]
\vspace*{-0.1in}
\centering
\input{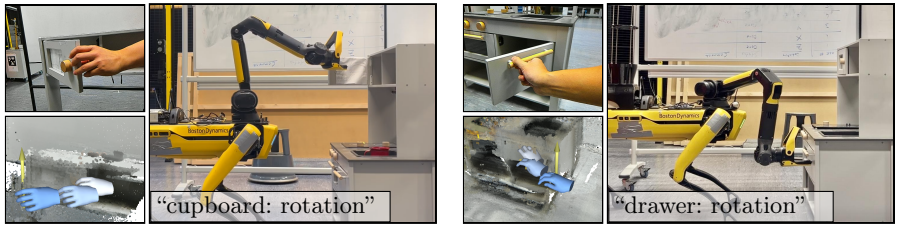}
\caption{\textbf{\ours~for robot manipulation.} \textit{Left:} Spot closes the cupboard. \textit{Right:} Spot opens the drawer. Insets show egocentric videos of hand-object interactions and the reconstructed 3D articulations.
\vspace*{-0.2in}
}
\label{fig:qualitative_app_robot}
\end{figure}

We validate the effectiveness of our approach in real-world settings using a Boston Dynamics Spot robot.
The robot is positioned in front of an IKEA toy kitchen and record the RGB-D egocentric video.
After processing human demonstration videos, our pipeline successfully extracts articulation parameters for the furniture, including cupboards and drawers.
We do not perform additional view selection on the input videos, as the human-recorded demonstrations exhibit limited viewpoint variation.
Given the extracted articulation model and estimated motion direction, the Spot robot infers (1) the opening orientation and (2) suitable contact points on the target object. This information is used to parameterize a Dynamic Movement Primitive (DMP)~\cite{ijspeert2013dynamical} that encodes the corresponding manipulation behavior. The robot then executes this motion primitive in an open-loop fashion to perform the desired opening or closing action, as illustrated in \cref{fig:qualitative_app_robot}.

\section{Conclusion}
We introduced \ours, a novel articulation extraction framework that jointly leverages hand–object interaction cues and geometric information from in-the-wild egocentric videos.
Our VLM-based reasoning pipeline exploits the commonsense knowledge embedded in foundation models to infer motion types, interacted objects, and plausible revolute joints.
To facilitate systematic benchmarking, we extend an existing in-the-wild egocentric video data set of real-world 3D indoor environments with 3D articulation annotations.
Extensive experiments on this data set and other public benchmarks demonstrate that our method consistently outperforms prior approaches in in-the-wild egocentric settings. We further show that the extracted articulations generalize well to downstream applications, including articulation prediction, and real-worldrobotic manipulation.
Overall, our results indicate that rich hand–object interaction information from large-scale egocentric videos can effectively support articulation understanding, without relying on high-fidelity 3D reconstruction or task-specific supervised training.

\section*{Acknowledgements}
AS acknowledges funding from the Research Council of Finland (grants 339730 and 362408). JK acknowledges funding from the Research Council of Finland (grants 352788, 353138, and 362407). We acknowledge CSC – IT Center for Science, Finland, and the Aalto Science-IT project for the computational resources. We thank Qilin Chen and Haishan Wang for helpful discussions.
\bibliographystyle{splncs04}
\bibliography{main}

\clearpage
\setcounter{page}{1}
\appendix

\renewcommand{\thetable}{A\arabic{table}}
\renewcommand{\thefigure}{A\arabic{figure}}

\section*{Appendices}
Contents

\vspace{0.5em}
{\small
\noindent\hyperref[supp:method]{\textbf{A\quad Additional Methods Description}}\dotfill\pageref{supp:method}\\[2pt]
\hspace*{2em}\hyperref[supp:notation]{A.1\enspace Notations}\dotfill\pageref{supp:notation}\\[1pt]
\hspace*{2em}\hyperref[supp:method:hand]{A.2\enspace Hand Trajectory Extraction}\dotfill\pageref{supp:method:hand}\\[1pt]
\hspace*{2em}\hyperref[supp:localization]{A.3\enspace Sparse Localization}\dotfill\pageref{supp:localization}\\[1pt]
\hspace*{2em}\hyperref[supp:geo]{A.4\enspace Geometry Reconstruction}\dotfill\pageref{supp:geo}\\[1pt]
\hspace*{2em}\hyperref[supp:manhattan]{A.5\enspace Manhattan Frame Extraction}\dotfill\pageref{supp:manhattan}\\[1pt]
\hspace*{2em}\hyperref[supp:method:vlm_settings]{A.6\enspace VLM Settings}\dotfill\pageref{supp:method:vlm_settings}\\[1pt]
\hspace*{2em}\hyperref[supp:method:final]{A.7\enspace Joint Inference}\dotfill\pageref{supp:method:final}\\[4pt]
\noindent\hyperref[supp:impl]{\textbf{B\quad Additional Experiment Details}}\dotfill\pageref{supp:impl}\\[2pt]
\hspace*{2em}\hyperref[supp:hyperparameters]{B.1\enspace Hyperparameters}\dotfill\pageref{supp:hyperparameters}\\[1pt]
\hspace*{2em}\hyperref[supp:annotation]{B.2\enspace Ground Truth Data Annotations}\dotfill\pageref{supp:annotation}\\[1pt]
\hspace*{2em}\hyperref[supp:benchmark]{B.3\enspace Benchmark List Selection}\dotfill\pageref{supp:benchmark}\\[1pt]
\hspace*{2em}\hyperref[supp:baselines]{B.4\enspace Adaptation of Baselines}\dotfill\pageref{supp:baselines}\\[4pt]
\noindent\hyperref[supp:exp]{\textbf{C\quad Additional Experimental Results}}\dotfill\pageref{supp:exp}\\[2pt]
\hspace*{2em}\hyperref[supp:qualitative]{C.1\enspace Qualitative Results}\dotfill\pageref{supp:qualitative}\\[1pt]
\hspace*{2em}\hyperref[supp:scene_agg]{C.2\enspace Scene-level Aggregation Results}\dotfill\pageref{supp:scene_agg}\\[4pt]
\noindent\hyperref[supp:limitations]{\textbf{D\quad Limitations and Failure Analysis}}\dotfill\pageref{supp:limitations}\\[0pt]
}
\vspace{1em}

\section{Additional Methods Description}
\label{supp:method}

\subsection{Notations}
\label{supp:notation}
For convenience, \cref{tab:notation} summarizes the key mathematical notation used throughout \ours.
\begin{table*}[ht!]
\centering
\caption{Summary of key mathematical notations used in \ours.}
\label{tab:notation}
\begin{tabular}{lp{9cm}}
\toprule
\textbf{Notation} & \textbf{Definition} \\
\midrule
\multicolumn{2}{c}{\textit{System Input \& Temporal Sets}} \\
\midrule
$\mathbf{V}$ & Egocentric RGB video ($\in\mathbb{R}^{T\times H\times W\times 3}$) with $T$ frames \\
$\mathcal{L}, l_j, M$ & Set of $M$ action descriptions; $l_j$ is a single action \\
$\mathbf{v}_j$ & Temporal clip for action $l_j$ \\
$K$ & Frames sparsely sampled from $\mathbf{v}_j$ \\
$K'$ & Frames with visible hands for MANO estimation \\
$N$ & Views sampled for geometry recovery \\
\midrule
\multicolumn{2}{c}{\textit{Dynamic Interaction Perception (Motion Analysis)}} \\
\midrule
$\mathbf{z}_t$ & Noisy fingertip observation at time $t$ \\
$\mathbf{x}_t$ & Hand kinematic state $[x_t, v^x_t, y_t, v^y_t, z_t, v^z_t]^\top \in \mathbb{R}^6$ \\
$\hat{\mathbf{p}}_t$ & RTS-smoothed hand position after contact filtering \\
$\mathcal{V}, v, n_v$ & Voxel grid, voxel $v$, and its multi-view count \\
$\mathcal{V}_{\text{conf}}$ & High-confidence voxels for coarse object localization \\
\midrule
\multicolumn{2}{c}{\textit{Static Scene Geometry \& Kinematic Priors}} \\
\midrule
$\mathcal{I}_s, I_n$ & $N$ representative views; $I_n$ is the $n$-th image \\
$\mathbf{K}, \mathbf{P}_n$ & Intrinsics and camera-to-world pose for $I_n$ \\
$\mathbf{D}_n, \mathbf{N}_n$ & Depth and normal maps for $I_n$ \\
$\mathcal{L}_{2D}, l_{2D}$ & Detected 2D line segments; $l_{2D}$ is one segment \\
$\mathcal{S}_{l_{2D}}, \mathbf{u}_m$ & Sampled pixels $\mathbf{u}_m{=}(u_m,v_m)$ on $l_{2D}$ \\
$\mathcal{P}_{l_{2D}}$ & Unprojected 3D points ($\subset\mathbb{R}^3$) for line $l_{2D}$ \\
$\mathcal{L}_{3D}, l_{3D}$ & Triangulated 3D line candidates; $l_{3D}$ is one line \\
$\{\mathbf{m}_1,\mathbf{m}_2,\mathbf{m}_3\}$ & Manhattan-world orthogonal directions (prismatic candidates) \\
\midrule
\multicolumn{2}{c}{\textit{Output Kinematics}} \\
\midrule
$\phi_{obj}, \phi_{scene}$ & Object- and scene-level articulation parameters \\
$\phi_i = \{c_i, a_i, o_i\}$ & Articulation parameters for joint $i$ \\
$c_i$ & Motion type ($\in\{\text{prismatic},\text{revolute}\}$) via VLM \\
$a_i$ & Motion axis ($\in\mathbb{R}^3$) \\
$o_i$ & Pivot point for revolute joints ($\in\mathbb{R}^3$) \\
\bottomrule
\end{tabular}

\end{table*}

\subsection{Hand Trajectory Extraction}

\label{supp:method:hand}
\subsubsection{Fingertip landmark aggregation}
As noted in \cref{sec:method:hand}, we represent the hand contact point $\mathbf{z}_t$ as the mean of the thumb, index, and middle fingertip landmarks in 3D world coordinates:
$\mathbf{z}_t = \tfrac{1}{3}(\mathbf{p}^\text{thumb}_t + \mathbf{p}^\text{index}_t + \mathbf{p}^\text{middle}_t)$.
This equal weighting is justified by grasp contact statistics~\cite{Brahmbhatt_2020_ECCV, GRAB:2020}, which report near-uniform contact likelihoods (${\approx}100\%$, $96\%$, $92\%$) for the thumb, index, and middle fingers, while ring and pinky fingers contribute substantially less.
An equal-weighted mean thus provides a stable, unbiased estimate of the grasp centroid without requiring force-based priors.

\subsubsection{Trajectory smoothing}
\begin{figure}[t!]
    \centering
    \includegraphics[width=\linewidth]{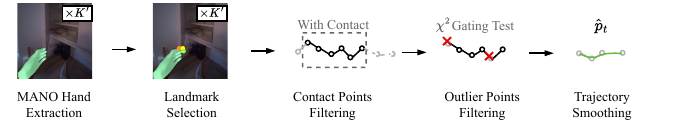}
    \caption{Illustration of the hand filtering pipeline. Starting from noisy MANO fingertip observations $\mathbf{z}_t$, we apply contact-based trimming, forward Kalman filtering with $\chi^2$ outlier rejection and RTS smoothing to obtain the refined trajectories $\{\hat{\mathbf{p}}_t\}$ used for articulation parameter estimation.}
    \label{fig:method:hand}
\end{figure}

After obtaining the hand trajectories $\mathbf{z}_t$, we leverage hand-object contact detection~\cite{cheng_towards_2023} to further remove trajectory segments corresponding to pre- and post-interaction motions. We then model them using a linear Gaussian state-space model with an integrated Ornstein-Uhlenbeck (OU) motion prior.
The latent kinematic state of the hand at time $t$ is defined as
\begin{equation}
\mathbf{x}_t = [x_t, v^x_t, y_t, v^y_t, z_t, v^z_t]^\top \in \mathbb{R}^6,
\end{equation}
where $(x_t, y_t, z_t)$ denotes the 3D position and $(v^x_t, v^y_t, v^z_t)$ denotes the corresponding velocity.
Following~\cite{anderson2015gp}, for each time interval $\Delta t$, the state dynamics are given by
\begin{equation}
\mathbf{x}_t = \mathbf{F}(\Delta t)\mathbf{x}_{t-1} + \boldsymbol{\epsilon}_t,
\end{equation}
where $\boldsymbol{\epsilon}_t \sim \mathcal{N}(\mathbf{0}, \mathbf{Q}(\Delta t))$, and the matrices $\mathbf{F}(\cdot)$ and $\mathbf{Q}(\cdot)$ are obtained from the exact discretization of the continuous-time OU process.
The noisy observations follow
\begin{equation}
\mathbf{z}_t = \mathbf{H}\mathbf{x}_t + \boldsymbol{\eta}_t,
\end{equation}
where $\mathbf{H}$ selects the position components and $\boldsymbol{\eta}_t \sim \mathcal{N}(\mathbf{0}, \sigma_{\mathrm{obs}}^2 \mathbf{I})$.
We perform a forward Kalman filtering pass~\cite{jouni2010kalman} and reject outliers using a $\chi^2$ gating test based on the squared Mahalanobis distance of the innovation.
Subsequently, we apply Rauch--Tung--Striebel (RTS) smoothing to obtain refined position estimates $\{\hat{\mathbf{p}}_t\}_{t=1}^{K'}$.
This leaves the effective smoothed articulation-related trajectories $\{\hat{\mathbf{p}}_t \in \mathbb{R}^3\}$, which are crucial for determining the final motion origin $o_i$.
A detailed illustration is shown in \cref{fig:method:hand}.

\subsection{Sparse Localization}
\label{supp:localization}
Each interaction clip $\mathbf{v}_j$ typically spans 30 to 100 frames, from which we select $5$ local frames that are inferred to contain active human–object interaction.
GroundedSAM is further used to filter out point cloud regions that do not belong to the interacted object.
The selected local frames are then used to identify high-confidence voxels (as defined in \cref{sec:method:hand}).
Next, we search over the full set of global camera frames and select 50 frames based on the visibility of the high-confidence voxels and furthest point sampling over camera positions (see \cref{sec:method:geo}).

\subsection{Geometry Reconstruction}
\label{supp:geo}
We recover a static scene reconstruction per data set.
For HD-EPIC~\cite{perrett_hd-epic_2025}, we run \textit{Map-Anything}~\cite{keetha2026mapanything} on a filtered subset of frames, excluding narrated interaction segments and frames with visible hands to retain only static observations.
Approximately 50-100 filtered frames with ground-truth camera poses are fed into the model to produce a dense 3D point cloud.
For Epic-Fields~\cite{EPICFields2023} and Arti4D~\cite{arti25werby}, we directly use the provided scene mesh or point cloud.
In all cases, the reconstruction is voxelized at $0.05\,\mathrm{m}$ resolution for downstream sparse localization.

\subsection{Manhattan Frame Extraction}
\label{supp:manhattan}
Indoor scenes exhibit strong structural regularity, with room layouts and furniture surfaces aligned to a few dominant orthogonal directions --- the Manhattan World Assumption~\cite{coughlan2000manhattan, furukawa2009manhattan}. Recent methods~\cite{guo2022manhattan, Popovic23ManhattanDF} extend this from room structures to object-level components such as cabinets and tables, providing natural priors for prismatic articulation axes.

Following~\cite{straub2017manhattan, Popovic23ManhattanDF}, we estimate a global Manhattan frame from the view set $\mathcal{I}_s$. For each frame, per-pixel surface normals and metric depth are computed using~\cite{wang2025moge2} and clustered via $k$-means. The centroid of the largest cluster gives the first dominant direction $\mathbf{m}_1$; two further orthogonal directions are obtained by solving:
\[
\min_{\mathbf{c}_s, \mathbf{c}_t} \; |\mathbf{c}_s^{\top}\mathbf{m}_1| + |\mathbf{m}_1^{\top}\mathbf{c}_t| + |\mathbf{c}_s^{\top}\mathbf{c}_t|,
\]
where $\mathbf{c}_s, \mathbf{c}_t$ are $k$-means cluster centroids. The per-frame axes are transformed to world coordinates via the camera pose and aggregated across all frames via mean-shift clustering, yielding the global orthogonal basis $\{\mathbf{m}_1, \mathbf{m}_2, \mathbf{m}_3\}$ used to constrain prismatic axis estimation (see \cref{sec:method:final}).

\subsection{VLM Settings}
\label{supp:method:vlm_settings}

\begin{figure}[ht!]
\centering
\begin{promptbox}
{\small
You are given a single frame from an egocentric (first-person) video clip.

\textbf{IMPORTANT CONTEXT:}
\begin{itemize}[leftmargin=1.3em,itemsep=1pt,topsep=2pt]
  \item This frame depicts an indoor kitchen/office scene captured from an egocentric (first-person) video.
  \item A human is interacting with a piece of furniture or fixture (such as a cupboard door, drawer, or fridge).
  \item \texttt{\{total\_frames\}} images are chosen from this video clip. This image is frame \texttt{\{frame\_idx\}/\{total\_frames\}}.
  \item The frame may correspond to one of three stages: before interaction (approaching), during interaction (motion occurring), or after interaction (motion completed).
\end{itemize}

\textbf{LANGUAGE AND PRIOR CUE:}
\begin{itemize}[leftmargin=1.3em,itemsep=1pt,topsep=2pt]
  \item The language description corresponding to this frame is: ``\texttt{\{narration\_text\}}''.
  \item Use the language description as a strong cue for the following task.
  \item The closest furniture detected from the hand trajectory in the 3D scene with semantic labels is: ``\texttt{\{fixture\}}''.
  \item The closest furniture may be incorrect due to noise or bias in the hand trajectory.
\end{itemize}

\textbf{TASK:}
\begin{enumerate}[leftmargin=1.7em,itemsep=1pt,topsep=2pt]
  \item Determine the furniture that the hand is interacting with.
  \item Determine the motion type of the furniture being interacted with in this frame.
\end{enumerate}

\textbf{GUIDELINES:}
\begin{itemize}[leftmargin=1.3em,itemsep=1pt,topsep=2pt]
  \item If the human is only approaching or the interaction has already finished and you cannot confidently determine the motion type, answer \texttt{unknown} for both the motion type and furniture name.
  \item You may infer the motion type from the language description, from hand movement, the relative position of the furniture parts, or contextual visual cues in the scene.
  \item Possible furniture names include: \texttt{cupboard, cabinet, drawer, fridge, refrigerator, oven, cooker, dishwasher, microwave, freezer, washmachine}. It may also be other furniture in the scene.
  \item Output the furniture name (not the part). For example, if interacting with a cabinet door, answer \texttt{cabinet} (not \texttt{door}).
  \item Motion type mapping (if visible / inferable):
  \begin{itemize}[leftmargin=1.4em,itemsep=1pt,topsep=1pt]
    \item Rotation $\rightarrow$ hinged motion (\eg, door, lid, or oven door).
    \item Translation $\rightarrow$ linear motion (\eg, sliding drawer).
  \end{itemize}
\end{itemize}
}
\end{promptbox}
\caption{Prompt for dynamic interaction perception.}
\label{fig:prompt_vqa_1}
\end{figure}

For both VLM tasks, we use OpenAI GPT-4o.
For dynamic interaction perception, we uniformly sample 10 frames from each interaction clip and present them to the VLM in temporal order.
The prompt for VLM-based motion reasoning is shown in \cref{fig:prompt_vqa_1} and \cref{fig:prompt_vqa_1b}.
For axis selection, we use the 50 global-view frames obtained from view reselection.
After multi-view line triangulation and grounding-mask filtering, we retain the four longest projected line segments and query the VLM using the prompt shown in \cref{fig:prompt_vqa_2}.

\begin{figure}[t]
\centering
\begin{promptbox}
{\small
\textbf{RESPONSE FORMAT:}
Respond with \textbf{only}:
\begin{center}
\texttt{furniture\_name:motion\_type}
\end{center}
\noindent where \texttt{motion\_type} is one of \texttt{rotation, translation, unknown}.

\textbf{EXAMPLES:}
\begin{itemize}[leftmargin=1.3em,itemsep=1pt,topsep=2pt]
  \item If the narration says ``open the cupboard door'', the closest fixture is \texttt{cupboard}, and the hand is near the handle $\rightarrow$ \texttt{cupboard:rotation}.
  \item If the narration says ``slide out the drawer'', and the drawer is partially extended $\rightarrow$ \texttt{drawer:translation}.
  \item If you cannot determine the furniture being interacted with $\rightarrow$ \texttt{unknown:unknown}.
  \item If you cannot determine the motion type $\rightarrow$ \texttt{cupboard:unknown} or \texttt{unknown:unknown}.
\end{itemize}
}
\end{promptbox}
\caption{Prompt for dynamic interaction perception (cont'd).}
\label{fig:prompt_vqa_1b}
\end{figure}

\begin{figure}[t]
\centering
\begin{promptbox}
{\small
You are given an indoor scene captured from an egocentric (first-person) video.
Your task is to identify the possible rotation axes of furniture parts based on the provided image.

\textbf{IMPORTANT CONTEXT:}
\begin{itemize}[leftmargin=1.3em,itemsep=1pt,topsep=2pt]
  \item The image shows a realistic indoor scene with various furniture and fixtures such as cupboards, drawers, ovens, or dishwashers.
  \item The image may contain four or fewer yellow line segments labeled with numbers \texttt{1}, \texttt{2}, \texttt{3}, and \texttt{4}.
  \item Each line is drawn in bold yellow with a white numeric label.
  \item These lines represent candidate axes; your goal is to determine which of them plausibly correspond to a \textbf{rotation axis} of a movable furniture part.
\end{itemize}

\textbf{TASK:}
\begin{enumerate}[leftmargin=1.7em,itemsep=1pt,topsep=2pt]
  \item Carefully examine the image and identify whether any numbered yellow line aligns with a hinge or rotation axis of a furniture part.
  \item Consider the furniture geometry and prior knowledge of common kitchen mechanisms (\eg, doors rotate, drawers translate).
  \item If none of the lines correspond to a rotation axis, answer \texttt{none}.
\end{enumerate}

\textbf{RESPONSE FORMAT:}
Please respond with one of the following:
\begin{itemize}[leftmargin=1.3em,itemsep=1pt,topsep=2pt]
  \item The line number(s) that represent rotation axes (\eg, \texttt{1} or \texttt{1, 3}) and the corresponding furniture part name.
  \item \texttt{none}, if no line represents a rotation axis.
\end{itemize}

\textbf{EXAMPLE RESPONSE:}
\begin{center}
\texttt{Line 2: cupboard door hinge}
\end{center}
}
\end{promptbox}
\caption{Prompt for selecting plausible rotation axes.}
\label{fig:prompt_vqa_2}
\end{figure}

\subsection{Joint Inference}
\label{supp:method:final}

Using the refined hand trajectories from \cref{sec:method:hand} and the geometric estimates from \cref{sec:method:geo}, we associate the smoothed hand trajectory $\{\hat{\mathbf{p}}_t\}_{t=1}^{K'}$ with the interacted object.
For \textbf{prismatic motion} ($c_i = \text{prismatic}$), the articulation axis $a_i$ is estimated by selecting the dominant Manhattan direction $\mathbf{m} \in \{\mathbf{m}_1, \mathbf{m}_2, \mathbf{m}_3\}$ that best aligns with the primary hand motion direction $\mathbf{v}_h$:
\begin{equation}
a_i = \arg\max_{\mathbf{m} \in \{\mathbf{m}_1,\mathbf{m}_2,\mathbf{m}_3\}} |\mathbf{v}_h^\top\mathbf{m}|.
\end{equation}
We set the motion origin $o_i$ for prismatic joints to the initial hand contact point $o_i = \hat{\mathbf{p}}_1$.
For \textbf{revolute motion} ($c_i = \text{revolute}$), we evaluate the semantically verified 3D axis proposals $l_{3D} \in \mathcal{L}_{3D}^*$ obtained from the VLM grounding step.
We parameterize each line $l_{3D}$ by a unit direction vector $\mathbf{u}_{l}$ and a point $\mathbf{q}_{l}$ on the line.
The orthogonal distance from a hand point $\hat{\mathbf{p}}_t$ to the line is computed as $d_t(l_{3D}) = \|(\hat{\mathbf{p}}_t - \mathbf{q}_{l}) \times \mathbf{u}_{l}\|$.
We select the optimal rigid axis $a_i$ by evaluating its geometric consistency with the dynamic hand trajectory, specifically by minimizing the variance of this distance to enforce a consistent rotation radius:
\begin{equation}
a_i = \arg\min_{l_{3D} \in \mathcal{L}_{3D}^*} \frac{1}{K'}\sum_{t=1}^{K'}\left(d_t(l_{3D}) - \bar{r}_l\right)^2,
\end{equation}
where $\bar{r}_l = \frac{1}{K'} \sum_{t=1}^{K'} d_t(l_{3D})$ represents the mean radius of rotation for that candidate axis.
The motion origin $o_i$ is defined as the 3D hand–object contact pivot.
We estimate it by computing the orthogonal projection of the hand trajectory centroid $\bar{\mathbf{p}} = \frac{1}{K'} \sum_{t=1}^{K'} \hat{\mathbf{p}}_t$ onto the selected axis $a_i$:
\begin{equation}
o_i = \mathbf{q}_{a_i} + \left((\bar{\mathbf{p}} - \mathbf{q}_{a_i})^\top \mathbf{u}_{a_i}\right)\mathbf{u}_{a_i}.
\end{equation}
This yields the final articulation parameters $\phi_i = \{ c_i, a_i, o_i \}$.

\section{Additional Experiment Details}
\label{supp:impl}

\subsection{Hyperparameters}
\label{supp:hyperparameters}

We summarize the key hyperparameters of the \ours~pipeline in \cref{tab:hyperparameters}.

\begin{table}[h]
\centering\small
\caption{Key hyperparameters of the \ours~pipeline.}
\label{tab:hyperparameters}
\vspace*{-.1in}
\begin{tabular}{llc}
\toprule
\textbf{Module} & \textbf{Parameter} & \textbf{Value} \\
\midrule
\multirow{4}{*}{\shortstack[l]{Hand Trajectory \\ Smoothing}}
  & Length scale $\ell$ & $10.0$\,m \\
  & Observation noise $\sigma_{\mathrm{obs}}$ & $0.05$\,m \\
  & Process noise std $q$ & $0.01$\,m \\
  & $\chi^2$ gating $p$-value ($\mathrm{df}{=}3$) & $0.05$ \\
\midrule
\multirow{4}{*}{\shortstack[l]{Scene Reconstruction \\ \& Localization}}
  & Voxel size & $0.05$\,m \\
  & Min.~views for high-confidence voxel & $4$ \\
  & Local interaction frames ($N_\ell$) & $5$ \\
  & Global views for geometry ($N_g$) & $50$ \\
\midrule
\multirow{4}{*}{VLM Queries}
  & Frames for motion type ($K$) & $10$ \\
  & Top candidate lines per view & $4$ \\
  & Temperature & $0.5$ \\
  & Top-$p$ & $0.3$ \\
\midrule
\multirow{4}{*}{\shortstack[l]{Revolute \\ Axis Fitting}}
  & Torus tolerance ratio & $0.15$ \\
  & Min.~torus tolerance & $0.015$\,m \\
  & Max.~torus tolerance & $0.050$\,m \\
  & Max.~rotation radius & $1.0$\,m \\
\midrule
\multirow{2}{*}{\shortstack[l]{Prismatic \\ Axis Fitting}}
  & Distance tolerance & $0.02$\,m \\
  & Min.~inlier rate & $0.3$ \\
\bottomrule
\end{tabular}
\end{table}

\subsection{Ground Truth Data Annotations}
\label{supp:annotation}

\begin{figure}
    \centering
    \input{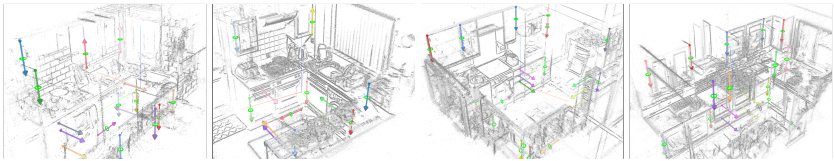}
    \caption{Visualization of annotated scenes in extended HD-EPIC \cite{perrett_hd-epic_2025} data set.}
    \label{fig:annotation_examples}
\end{figure}

To evaluate articulation perception accuracy, we annotate ground-truth articulations using a custom annotation tool built on Open3D for interactive point selection.
We initialize the articulation origin by selecting points from the reconstructed SLAM point cloud, and specify the articulation axis direction using mesh vertices.
Manual corrections are applied in post-processing to ensure annotation accuracy.
Annotated examples are shown in \cref{fig:annotation_examples}.

\subsection{Benchmark List Selection}
\label{supp:benchmark}

From the full HD-EPIC~\cite{perrett_hd-epic_2025} and Epic-Fields~\cite{EPICFields2023} data sets, we construct \textbf{EgoArti}, a benchmark of over 300 interaction clips spanning more than 50 articulated object categories.
Clips are selected based on their language descriptions: we retain narrations containing an interactable furniture noun and an articulation verb (\eg, ``open'' or ``close'').
An example of the benchmark list is shown in \cref{fig:supp:benchmark}. The full list is included in our supplementary files.

\begin{figure}[h]
\centering
\begin{promptbox}[width=0.6\linewidth]
{\ttfamily\small
{
  "P01-20240202-110250-1":   "cupboard.009",\\
  "P01-20240202-110250-60":  "drawer.006",\\
  "P01-20240202-161948-8":   "drawer.003",\\
  "P01-20240202-171220-115": "cupboard.008",\\
  "P01-20240202-195538-209": "cupboard.008",\\
  "P01-20240203-150506-24":  "oven.001"
}
}
\end{promptbox}
\caption{An example of the benchmark list.}
\label{fig:supp:benchmark}
\end{figure}

\subsection{Adaptation of Baselines}
\label{supp:baselines}

\subsubsection{Articulation3D~\cite{Qian22}}
Articulation3D is built upon Mask R-CNN; we follow the official implementation and resize all inputs to $640\times480$ to preserve its original performance. The pipeline predicts a 2D plane mask, a 2D articulation axis, and 3D plane parameters (normal and offset). In our adaptation, we replace the plane-parameter prediction head with metric depth from~\cite{wang2025moge2}, as accurate depth is required to lift 2D predictions into 3D space. We further incorporate ground-truth camera poses into the temporal optimization, modifying the objective to:
$$r(\alpha,t)=\mathrm{IoU}\!\left(\mathcal{M}^{(t)},\,\mathbf{K}\mathbf{T}_{tgt}^{-1}\mathbf{T}_{ref}\,[\mathbf{R}_{\alpha},\mathbf{t}_{\alpha}]\,\Pi\right),$$
which accounts for camera motion between frames.

\subsubsection{ArtiPoint~\cite{arti25werby}}
The original ArtiPoint enforces a minimum interaction duration of 30 frames (one second at 30\,fps) to filter spurious hand detections. However, many clips in HD-EPIC~\cite{perrett_hd-epic_2025} involve significantly shorter interactions, causing a large fraction of valid clips to be discarded under the default setting. We therefore relax this temporal constraint and treat an interaction as valid upon any hand detection, allowing the method to operate on the short, in-the-wild interactions present in our benchmark.

\subsubsection{Articulate-Anything (AA)~\cite{le2024articulate}}
Articulate-Anything is an object-level articulated reconstruction and generation method. For in-the-wild inputs, it detects the object category via a VLM and retrieves a matching mesh from the PartNet-Mobility data set~\cite{sapien}. Since the original implementation does not align the retrieved object's articulation to world coordinates, a direct quantitative comparison is not feasible. We therefore include only qualitative results for this baseline.

\subsubsection{iTACO~\cite{peng_itaco_2025}}
We adapt iTACO to use ground-truth camera poses, as the object segmentation mask is often unreliable in our setting, which degrades the downstream LoFTR-based pose estimation used in iTACO's coarse joint estimation stage. Additionally, iTACO's AutoSeg-SAM2~\cite{AutoSeg_SAM2} segmentation performs best when the video begins from an open state; under in-the-wild conditions where the interaction includes both opening and closing phases, this assumption frequently fails. We therefore provide only the opening phase of each interaction for iTACO inference in our qualitative evaluation. For prismatic joints, iTACO outputs only the motion direction; we use the ground-truth motion origin as the start position for visualization. Finally, as iTACO relies on a scanned mesh for 3D digital twin reconstruction, which is unavailable in in-the-wild settings, we report only the estimated joint parameters.

\section{Additional Experimental Results}
\label{supp:exp}

\subsection{Qualitative Results}
\label{supp:qualitative}
In this section, we provide additional qualitative results on the HD-EPIC \cite{perrett_hd-epic_2025} and Arti4D \cite{arti25werby} data sets (see Figure \ref{fig:qualitative_1}). Specifically, the first two columns ("cupboard" and "drawer") feature examples from the HD-EPIC data set, while the latter two columns ("microwave" and "cabinet") feature examples from the Arti4D data set.

\subsubsection{Articulate-Anything \cite{le2024articulate}:} This method generally performs well for object simulation and accurately identifies the interacted object in most scenarios.
However, it is sensitive to lighting conditions; for example, it misclassifies the object in the dark "drawer" sequence as a box (Fig. \ref{fig:qualitative_1}).
Furthermore, its data set retrieval approach fundamentally limits its ability to recover accurate, instance-specific dimensions.

\subsubsection{ArtiPoint \cite{arti25werby}:} This baseline struggles on the HD-EPIC \cite{perrett_hd-epic_2025} data set due to depth estimation errors, a lack of robust tracking points, and heavy occlusions of openable parts.
As shown in the "cupboard" and "cabinet" cases, it tracks incorrect keypoints, leading to inaccurate articulation extraction. Even when tracking succeeds, such as in the "drawer" case, underlying depth errors compromise the final graph-based extraction.

\subsubsection{iTACO \cite{peng_itaco_2025}:} The performance of iTACO is bottlenecked by its heavy reliance on pre-processing, and therefore failed in our qualitative results. Specifically, MonST3R \cite{zhang_monst3r_2024} frequently fails to cleanly segment dynamic and static masks in our real-world settings, resulting in less plausible articulations.

\subsubsection{Ours:} In contrast, our approach demonstrates more consistent performance across challenging real-world scenarios. By effectively utilizing hand trajectory cues, our method accurately captures the interaction intent. While minor deviations exist (\eg a slight triangulation error in the "cabinet" sequence) our extracted articulations are visibly more accurate than those of the baselines and are directly suitable for downstream simulation.

\subsection{Scene-level Aggregation Results}
\label{supp:scene_agg}
\begin{figure}[ht]
    \centering
    \includegraphics[width=0.6\linewidth]{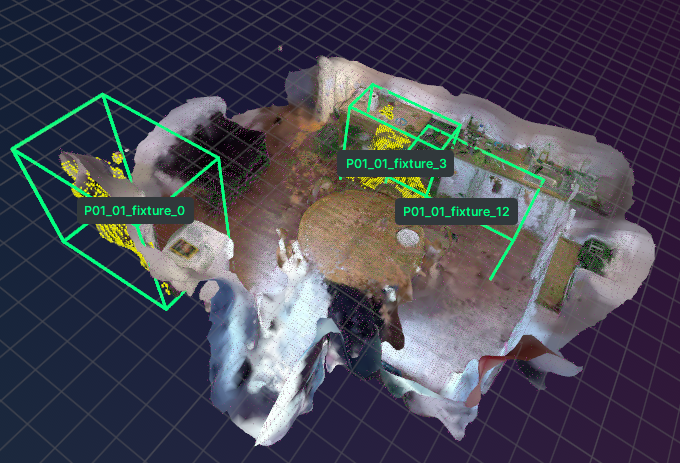}
    \caption{\textbf{Scene-level aggregation.} Multiple articulated object instances (highlighted by bounding boxes) are localized and annotated within the same reconstructed 3D scene, enabling scene-level aggregation of articulation estimates across repeated interactions.}
    \label{fig:scene_aggregation}
\end{figure}

In the HD-EPIC \cite{perrett_hd-epic_2025} and Epic-Fields \cite{EPICFields2023} datasets, the same interaction (\eg, ``open the cupboard A'') may occur multiple times. This allows us to aggregate multiple hand trajectories to obtain a more accurate estimate of the articulation.
For the HD-EPIC dataset, we group video clips corresponding to the same object based on the distance between the hand trajectory and the object mesh.
For Epic-Fields, we associate clips by computing the IoU between high-confidence voxelizations.
A visualization of this aggregation is shown in \cref{fig:scene_aggregation}.

\input{fig/4_qualitative_1}

\section{Limitations and Failure Analysis}
\label{supp:limitations}
\begin{figure*}[ht!]
    \centering\scriptsize
    \setlength{\figurewidth}{0.22\textwidth}
    \begin{tikzpicture}[
        image/.style = {inner sep=0pt, outer sep=1pt, minimum width=\figurewidth, anchor=north west, text width=\figurewidth},
        node distance = 1pt and 1pt, every node/.style={font={\tiny}},
        label/.style = {font={\scriptsize\bf\strut},anchor=north,inner sep=2pt},
        rowlabel/.style = {font={\scriptsize\bf\strut},anchor=east,inner sep=2pt},
        ]

        \node [image] (img-00) {\includegraphics[width=\figurewidth]{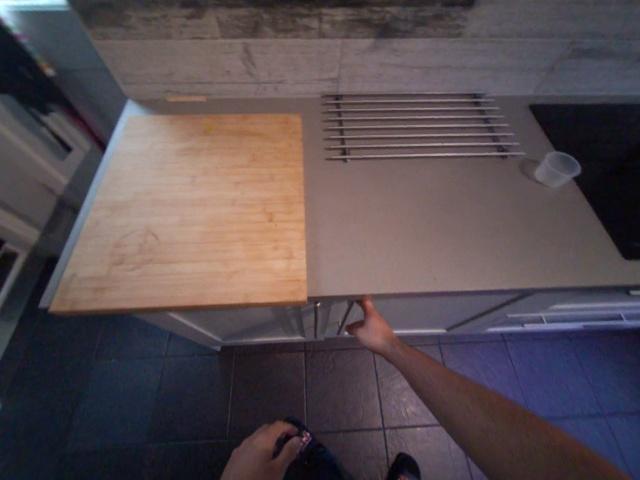}};
        \node [image,right=of img-00] (img-01) {\includegraphics[width=\figurewidth]{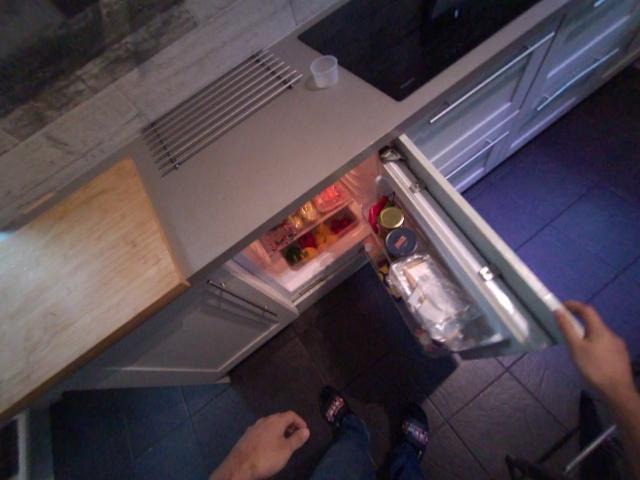}};
        \node [image,right=of img-01] (img-02) {\includegraphics[width=\figurewidth]{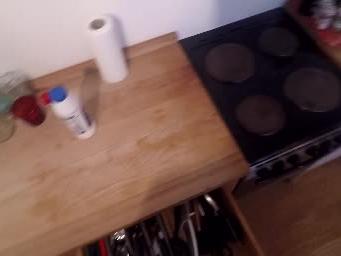}};
        \node [image,right=of img-02] (img-03) {\includegraphics[width=\figurewidth]{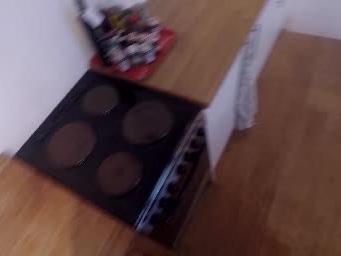}};

        \node[label] at (img-00.south) {(a)};
        \node[label] at (img-01.south) {(b)};
        \node[label] at (img-02.south) {(c)};
        \node[label] at (img-03.south) {(d)};

    \end{tikzpicture}
    \caption{\textbf{Failure Case Examples.}
    (a, b) A fridge may be misclassified as a ``cupboard'' due to its visually ambiguous exterior, causing both VLM reasoning and grounding to fail. (c, d) Examples where the interacting hand does not appear in the frame throughout the entire sequence, leading to tracking failures.
    }
    \label{fig:failure}
\end{figure*}

Our framework has two limitations.
First, articulation inference depends on the reliability of the underlying foundation models. Visually ambiguous objects, such as a fridge that resembles a cupboard when closed (\cref{fig:failure}(a, b)), can cause both the VLM and the grounding model to misidentify the object category, leading to incorrect axis estimation.
Second, under in-the-wild conditions, the hand may appear in very few frames during an interaction, providing insufficient trajectory points for robust articulation inference. This is especially common in the 'close' action, because humans tend to shift their gaze to the next action when closing the previous object. (\cref{fig:failure}(c, d)). This is a common challenge for all trajectory-based methods.
We leave addressing these limitations as future work.

\end{document}